\documentclass{article}
\usepackage{ijcai24}

\usepackage{times}
\usepackage{soul}
\usepackage{url}
\usepackage[hidelinks]{hyperref}
\usepackage[utf8]{inputenc}
\usepackage[small]{caption}
\usepackage{graphicx}
\usepackage{amsmath}
\usepackage{amsthm}
\usepackage{booktabs}
\usepackage{algorithm}
\usepackage{algorithmic}
\usepackage[switch]{lineno}

\usepackage{amsfonts,amssymb}
\usepackage{multirow}
\newcommand{\model}{FastKGE}
\newcommand{\mec}{IncLoRA}
\usepackage{color}

\title{Efficient Continual Knowledge Graph Embedding via Multi-Layer Incremental Low-Rank Learning}
\title{Efficient Continual Knowledge Graph Embedding via Incremental Low-Rank Adapters}
\title{Fast and Continual Knowledge Graph Embedding via Incremental LoRA}

\author{
    Jiajun Liu$^1$
    \and
    Wenjun Ke$^{1,2}$\thanks{Corresponding author.} \and
    Peng Wang$^{1,2}$\footnotemark[1] \and
    Jiahao Wang$^1$\and
    Jinhua Gao$^3$\and \\
    Ziyu Shang$^1$\and
    Guozheng Li$^1$ \and
    Zijie Xu$^1$ \and 
    Ke Ji$^1$ \And
    Yining Li$^4$ 
    \\
    \affiliations
    $^1$School of Computer Science and Engineering, Southeast University\\
    $^2$Key Laboratory of New Generation Artificial Intelligence Technology and Its \\
    Interdisciplinary Applications (Southeast University), Ministry of Education \\
    $^3$Institute of Computing Technology, Chinese Academy of Sciences\\
    $^4$College of Software Engineering, Southeast University\\
    \emails
    \{jiajliu, kewenjun, pwang, wang\_jh, ziyus1999, gzli, zijiexu, keji, liyining\}@seu.edu.cn, \{gaojinhua\}@ict.ac.cn
}

\begin{document}

\maketitle

\begin{abstract}
Continual Knowledge Graph Embedding (CKGE) aims to efficiently learn new knowledge and simultaneously preserve old knowledge. 
Dominant approaches primarily focus on alleviating catastrophic forgetting of old knowledge but neglect efficient learning for the emergence of new knowledge. 
However, in real-world scenarios, knowledge graphs (KGs) are continuously growing, which brings a significant challenge to fine-tuning KGE models efficiently. 
To address this issue, we propose a fast CKGE framework (\model), incorporating an incremental low-rank adapter (\mec) mechanism to efficiently acquire new knowledge while preserving old knowledge. 
Specifically, to mitigate catastrophic forgetting, \model\ isolates and allocates new knowledge to specific layers based on the fine-grained influence between old and new KGs. 
Subsequently, to accelerate fine-tuning, \model\ devises an efficient \mec\ mechanism, which embeds the specific layers into incremental low-rank adapters with fewer training parameters. 
Moreover, \mec\ introduces adaptive rank allocation, which makes the LoRA aware of the importance of entities and adjusts its rank scale adaptively. 
We conduct experiments on four public datasets and two new datasets with a larger initial scale. 
Experimental results demonstrate that \model\ can reduce training time by 34\%-49\% while still achieving competitive link prediction performance against state-of-the-art models on four public datasets (average MRR score of 21.0\% vs. 21.1\%).
Meanwhile, on two newly constructed datasets, \model\ saves 51\%-68\% training time and improves link prediction performance by 1.5\%. 
\end{abstract}

\section{Introduction}
Knowledge graph embedding (KGE)~\cite{wang2017knowledge,rossi2021knowledge} aims to embed entities and relations in knowledge graphs (KGs)~\cite{dong2014knowledge} into vectors, which is crucial for various downstream applications, such as question answering~\cite{bordes2014open}, recommendation systems~\cite{zhang2016collaborative}, fact detecting~\cite{shang2024ontofact}, and information extraction~\cite{li2022fastre,xu2023two,li2024unlocking}. 
Traditional KGE methods~\cite{bordes2013translating,sun2019rotate,liu-etal-2020-aprile,pan-wang-2021-hyperbolic-hierarchy,shang2023askrl,liu2023iterde} primarily focus on dealing with static KGs. 
However, real-world KGs, such as Wikidata~\cite{vrandevcic2014wikidata} and YAGO~\cite{10.1145/1242572.1242667}, are dynamic and continuously evolving with emerging knowledge. 
For instance, Wikidata expanded from 16M entities to 46M between 2014 and 2018~\cite{wikidata_web}. 
A challenge with traditional KGE is that updating the embeddings of entities and relations requires retraining the entire KG, leading to heavy training costs in large-scale KGs. 
To address this issue, the continual knowledge graph embedding (CKGE) task has received growing attention to fine-tuning with only new knowledge~\cite{song2018enriching,daruna2021continual}.

\begin{figure}[t]
\centering
\includegraphics[width=0.48\textwidth]{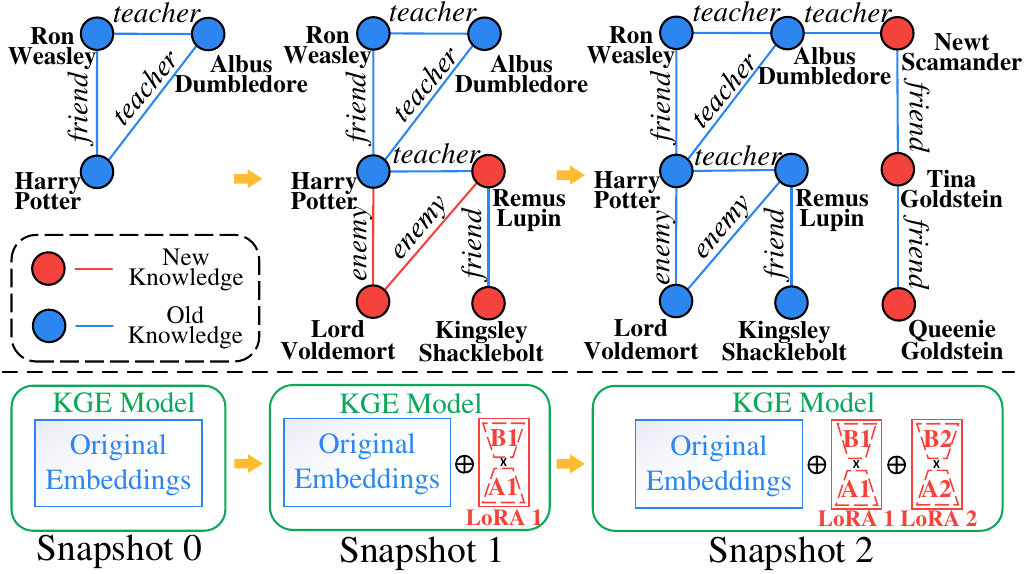} 
\caption{Illustration of IncLoRA for CKGE. The above is a growing KG about \textit{Harry Potter} as the storyline unfolds, and below are the KGE models with incremental LoRAs in each snapshot.}
\label{intro}
\end{figure}

The primary challenge in CKGE lies in alleviating catastrophic forgetting~\cite{kirkpatrick2017overcoming,liu2024towards} while simultaneously reducing training costs. 
One solution for CKGE, known as the full-parameter fine-tuning paradigm, memorizes old knowledge by replaying a core old data set~\cite{lopez2017gradient,wang2019sentence,kou2020disentangle} or introducing additional regularization constraints~\cite{zenke2017continual,kirkpatrick2017overcoming,cui2023lifelong}. 
Although this paradigm effectively mitigates catastrophic forgetting, it significantly increases training costs, especially when handling large-scale KGs. 
Another solution adopts the incremental-parameter fine-tuning paradigm, with only a few parameters to learn emerging knowledge~\cite{rusu2016progressive,lomonaco2017core50}. 
Despite eliminating explicit knowledge replay, the straightforward alignment of new and old parameter dimensions may still result in an unacceptable increase in parameters and training time. 
Meanwhile, in recent years, to efficiently reduce the training time of large language models (LLMs)~\cite{radford2019language,brown2020language} for downstream tasks, some work has employed low-rank adapters, such as LoRA~\cite{hu2022lora}, to lower the parameter dimension and enable efficient parameter fine-tuning. 
Namely, LoRA freezes the pre-trained model weights and injects trainable rank decomposition matrices into original model architectures. 
In this paper, we try to innovatively adapt this mechanism to address continual learning problems, i.e., CKGE.

Inspired by low-rank adaptation~\cite{hu2022lora} in parameter fine-tuning for LLMs, we are among the first to store new knowledge in KGs via low-rank adapters (LoRAs) to reduce training costs. 
As shown in Figure~\ref{intro}, in contrast to the old entity \textit{Ron Weasley}, new incremental LoRAs should be assigned to the new entity \textit{Remus Lupin}. 
Moreover, recent work has discovered that new parameters of different layers should be adaptively assigned distinct dimensions~\cite{ansuini2019intrinsic}, which can also be applied to emerging knowledge with different structural features in CKGE. 
As depicted in Figure~\ref{intro}, \textit{Remus Lupin} connects with more entities than \textit{Lord Voldemort}. 
Consequently, it retains a broader influence, requiring more parameters for effective learning.
In light of this consideration, we propose a fast CKGE framework (\model), incorporating a novel incremental low-rank adapter (\mec) mechanism, to learn new knowledge both effectively and efficiently. 
 To alleviate catastrophic forgetting, \model\ isolates new knowledge to specific layers. 
Concretely, \model\ sorts and divides the new entity representations into explicit layers according to their distance from the old KG and the degree centrality. 
 To reduce training costs, \mec\ embeds the specific layers into incremental low-rank adapters with fewer training parameters. 
Specifically, the rank scales of each layer are allocated adaptively with the awareness of the KG structure.

To make up for the deficiency of small initial KG size in current CKGE datasets~\cite{hamaguchi2017knowledge,kou2020disentangle,daruna2021continual,cui2023lifelong}, we construct two new CKGE datasets: FB-CKGE and WN-CKGE based on FB15K-237~\cite{dettmers2018convolutional} and WN-18RR~\cite{toutanova2015representing}. 
FB-CKGE and WN-CKGE allocate 60\% of total triples to generate the larger initial KG. 
Results on four traditional datasets show that \model\ reduces training time by 34\%-49\% while still achieving competitive MRR scores in link prediction tasks against SOTAs (average in 21.0\% vs. 21.1\%). 
Meanwhile, on the new datasets FB-CKGE and WN-CKGE, \model\ reduces 51\%-68\% training time while improving link prediction performance by 1.5\% in MRR on average.

To sum up, the contributions of this paper are three-fold: 
\begin{itemize}

    \item To the best of our knowledge, we are among the first to introduce low-rank adapters to CKGE, namely, emerging knowledge can be stored in low-rank adapters to reduce training costs and preserve old knowledge well. 

    \item We devise a fast CKGE framework (\model), which isolates knowledge into specific layers to alleviate catastrophic forgetting and utilizes an incremental low-rank adapter (\mec) mechanism to reduce training costs. 

    \item Experimental results demonstrate that \model\ significantly reduces training time with competitive performance in link prediction tasks compared with SOTAs. Additionally, two new open CKGE datasets with large-scale initial KGs are released. 

\end{itemize}

\begin{figure*}[t]
\centering
\includegraphics[width=1\textwidth]{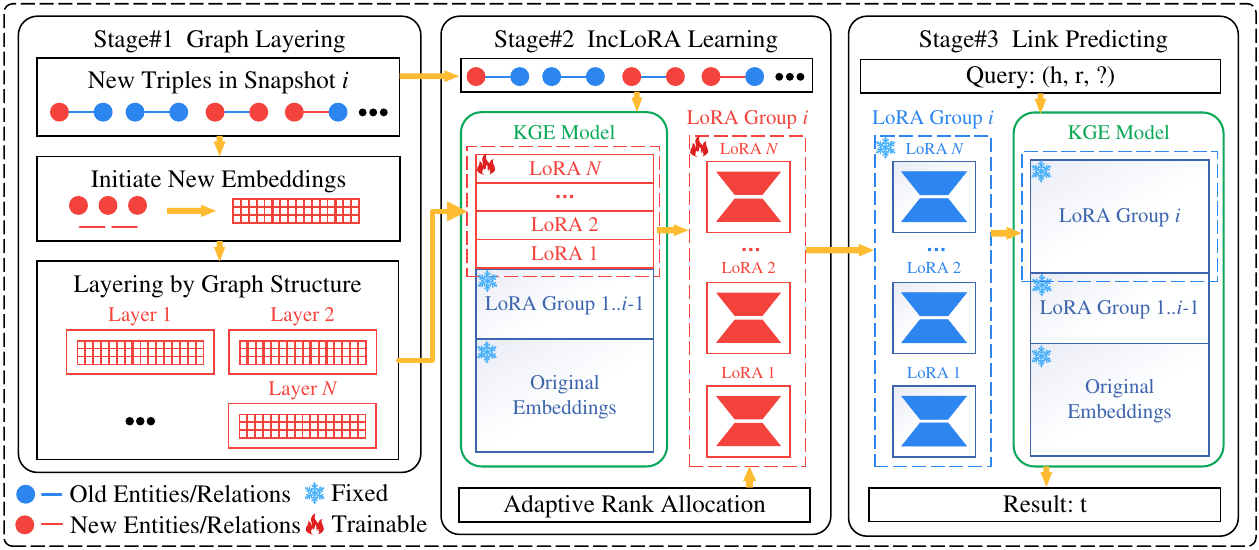} 
\caption{An overview of \model\ framework. LoRA group $i$ denotes the set for all LoRAs in a snapshot $i$.}
\label{framework_picture}
\end{figure*}

\section{Related Work}
In contrast to traditional knowledge graph embedding (KGE) approaches~\cite{bordes2013translating,trouillon2016complex,kazemi2018simple}, continual knowledge graph embedding (CKGE) methods~\cite{song2018enriching,daruna2021continual} enable KGE models to acquire new knowledge while retaining previously learned knowledge. 
Existing CKGE methods can be categorized into two main groups. 
First, full-parameter fine-tuning methods preserve learned knowledge by replaying old data~\cite{lopez2017gradient,wang2019sentence,kou2020disentangle}, or introducing constraints on weight updates in neural networks~\cite{zenke2017continual,kirkpatrick2017overcoming,cui2023lifelong}. 
Second, incremental-parameter fine-tuning methods~\cite{rusu2016progressive,lomonaco2017core50} adaptively adjust architectural properties to accommodate new information while preserving old parameters, thus facilitating continual learning. 
However, these methods only focus on preserving knowledge while ignoring training efficiency when KGs evolve. 

In the field of large language models (LLMs), some work tries to utilize low-rank adapters (LoRAs) to fine-tune LLMs efficiently~\cite{hu2022lora,zhang2023adaptive}. 
Based on it, recent work tries to utilize LoRAs to separately store knowledge to alleviate catastrophic forgetting in continual learning~\cite{wang2023orthogonal}. 
In the field of KGE, recent work tries to learn a small set of reserved entities to represent all entities for parameter-efficient learning~\cite{chen2023entity}. 
However, few works focus on efficient fine-tuning for CKGE.

\section{Methodology}
\subsection{Preliminary and Problem Statement}
\paragraph{Growing Knowledge Graph.}
A growing knowledge graph (KG) is denoted as snapshot sequences, i.e., $\mathcal{G}=\{\mathcal{S}_{0}, \mathcal{S}_{1}, ..., \mathcal{S}_{n}\}$. 
Snapshot $\mathcal{S}_{i}$ is a triplet $(\mathcal{E}_{i}, \mathcal{R}_{i}, \mathcal{T}_{i})$, where $\mathcal{E}_{i}$, $\mathcal{R}_{i}$ and $\mathcal{T}_{i}$ denote the set of entities, relations, and triples at time $i$, respectively. 
Furthermore, we denote $\Delta \mathcal{T}_{i} = \mathcal{T}_{i} - \mathcal{T}_{i - 1}$, $\Delta \mathcal{E}_{i} = \mathcal{E}_{i} - \mathcal{E}_{i - 1}$ and $\Delta \mathcal{R}_{i} = \mathcal{R}_{i} - \mathcal{R}_{i - 1}$ as new triples, entities, and relations, respectively.

\paragraph{Continual Knowledge Graph Embedding.}
Continual knowledge graph embedding (CKGE) aims to embed entities and relations into vectors in the growing KG $\mathcal{G}=\{\mathcal{S}_{0}, \mathcal{S}_{1}, ..., \mathcal{S}_{n}\}$. 
Specifically, when new triples $\Delta \mathcal{T}_{i}$ emerge in time $i$, CKGE learns the representations of new entities $\Delta \mathcal{E}_{i}$ and relations $\Delta \mathcal{R}_{i}$, and updates the representations of old entities $\mathcal{E}_{i-1}$ and relations $\mathcal{R}_{i-1}$ to adapt $\Delta \mathcal{T}_{i}$. 
Finally, all representations of entities $\mathcal{E}_{i}$ and relations $\mathcal{R}_{i}$ are obtained.

\subsection{Framework}
The framework of \model\ is illustrated in Figure~\ref{framework_picture}. 
Overall, as a KG grows in each snapshot, we utilize incremental low-rank adapters (LoRAs) for different KG layers to learn and preserve new entities and relations. 
First, during the knowledge graph layering stage, we divide new entities and relations into several layers based on their distances from the old graph and node degrees. 
Second, in the \mec\ learning stage, embeddings of entities and relations in each layer are represented by incremental LoRAs with adaptive rank allocation. 
Finally, in the link predicting stage, we compose all new LoRAs into a LoRA group and concat all LoRA groups and initial embeddings for inference.

\subsection{Graph Layering}
In order to achieve separate storage of KGs and assign different ranks to LoRAs of different layers, we divide new knowledge into several layers according to a graph structure. 
For new triples $\Delta \mathcal{T}_{i}$ emerging in a new snapshot $i>0$, we firstly get new entities $\Delta \mathcal{E}_{i}$ and relations $\Delta \mathcal{R}_{i}$ and initiate embeddings of them as shown in Stage 1 of Figure~\ref{framework_picture}. 
Then, to sort and divide embeddings of $\Delta \mathcal{E}_{i}$ in snapshot $i$, we calculate the importance of $\Delta \mathcal{E}_{i}$ by the distance from the old graph and degree centrality. 
Specifically, we use the breadth-first search (BFS) algorithm to progressively expand $\Delta \mathcal{E}_{i}$ in snapshot $i$ from $S_{i-1}$. 
Then, we get the sorted entity sequence $\mathbf{s}_{entity}$ as follows: 
\begin{equation}
    \mathbf{s}_{entity} = [e_{1}, e_{2}, ... , e_{|\Delta \mathcal{E}_{i}|}]
    \label{s_entity}
\end{equation}
where for $e_{j}, e_{k} \in \mathbf{s}_{entity}$, if $j \leq k$, the distance of $e_{j}$ from the old graph is closer than $e_{k}$. 
To further sort entities with the same distance, we denote $f_{dc}(e)$ as the degree centrality of $e$ in the new graph composed of new triples $\Delta \mathcal{T}_{i}$ as follows: 
\begin{equation}
    f_{dc}(e) = \frac{f_{neighbor}(e)}{|\Delta \mathcal{E}_{i}| - 1}
    \label{dc}
\end{equation}
where $f_{neighbor}(e)$ denotes the number of the neighbors of $e$ in $\Delta \mathcal{T}_{i}$. 
For entities with the same distance from the old graph in $\mathbf{s}_{entity}$, we use $f_{dc}$ for further sorting. 
Then, we divide the entities in $\mathbf{s}_{entitiy}$, previously sorted with importance, equally into $N$ distinct layers $\overline{E} = \{{E}_{1}, {E}_{2}, ..., {E}_{N}\}$, where $E_{k}$ denotes the $k$-th layer of entities, and $N$ is a hyper-parameter. 
For relations, we only put all new relations into a layer $\overline{R}$ rather than layering. 
That is because the number of entities is increasing more significantly than relations in the dynamic evolution of KGs, and the number of total embedding parameters linearly to the number of entities in practice~\cite{chen2023entity}. 
Therefore, we focus on storage and training optimization for new entities. 
Finally, we get the layering entities and relations $\overline{E}$ and $\overline{R}$.

\subsection{IncLoRA Learning}
\subsubsection{Incremental Low-Rank Decomposition}
To accelerate learning, we propose an incremental low-rank adapter learning mechanism (IncLoRA) to reduce training costs. 
Specifically, we obtain entity and relation layers $\overline{E}$ and $\overline{R}$ from graph layering stage. 
Then, for each layer in $\overline{E}$ and $\overline{R}$, we use an incremental low-rank adapter to learn and store knowledge. 
Take $\overline{E}$ as example, the embeddings of $k$-th layer $E_{k}$ in $\overline{E}$ can be denoted as $\mathbf{E}_{k}$. 
For learning the embeddings $\mathbf{E}_{k} \in \mathbb{R}^{n \times d}$, we learn matrices $\mathbf{A}_{k} \in \mathbb{R}^{n \times r}$ and $\mathbf{B}_{k} \in \mathbb{R}^{r \times d}$, making it meet the following condition:
\begin{equation}
    \mathbf{E}_{k} = \mathbf{A}_{k} \mathbf{B}_{k}
    \label{lora_composition}
\end{equation}
where $n$ denotes the number of entities in a layer $E_{k}$, $d$ stands for embedding dimension and $r$ refers to the rank of  $(\mathbf{A}_{k}, \mathbf{B}_{k})$. 
We denote $(\mathbf{A}_{k}, \mathbf{B}_{k})$ as an incremental LoRA. 
To make sure that parameters of low-rank learning is fewer than normal training for faster learning, the decomposition should meet the following condition: 
\begin{eqnarray}
    \label{condition_2}
    r &\leq& \frac{n \times d}{n + d}
\end{eqnarray}
By this way, we compose $\mathbf{E}_{k}$ to an incremental LoRA $(\mathbf{A}_{k}, \mathbf{B}_{k})$. 
Finally, we get a LoRA group $\mathbf{G}_{i}^{\mathbf{E}}$ for all embeddings of $\Delta \mathcal{E}_{i}$ as follows: 
\begin{equation}
    \mathbf{G}_{i}^{\mathbf{E}} = {\rm{concat}}(\{(\mathbf{A}_{k} \times \mathbf{B}_{k}) | 1 \leq k \leq N\})
    \label{loragroup}
\end{equation}
where concat($\cdot$) denotes the concatenation of several matrices. 
We can also get the LoRA group $\mathbf{G}_{i}^{\mathbf{R}}$ for new relations by the same decomposition.

\subsubsection{Adaptive Rank Allocation}
In order to preserve more information for more important entities, we utilize an adaptive rank allocation strategy to assign different ranks to LoRAs in different layers. 
Specifically, instead of assigning a fixed base rank $r_{base}$ to all incremental LoRAs in $\mathbf{G}_{i}^{\mathbf{E}}$, we assign more ranks to more important low-rank LoRAs with higher $f_{dc}$. 
Firstly, we denote the total sum of degree centrality in the $k$-th layer $Sum_{k}$ and the average degree centrality $Avg_{dc}$ in all layers as follows: 
\begin{eqnarray}
    \label{sum_dc}
    Sum_{k} &=& \sum_{e \in E_{k}} f_{dc}(e) \\
    \label{avg_dc}
    Avg_{dc} &=& \frac{\sum_{k \in N}Sum_{k}}{N}
\end{eqnarray}
Then, the $r_{k}$ of the $k$-th LoRA in $\mathbf{G}_{i}^{\mathbf{E}}$ is denoted as follows: 
\begin{eqnarray}
    \label{r_{k}}
    r_{k} &=& \frac{r_{base} \cdot Sum_{k}}{Avg_{dc}}
\end{eqnarray}
Finally, the rank of all adapters in $\mathbf{G}_{i}^{\mathbf{E}}$ is determined. 
Since there is only one layer in $\overline{R}$, we do not utilize adaptive rank allocation for $\mathbf{G}_{i}^{\mathbf{R}}$.

\subsubsection{IncLoRA Training}
From above, we obtain the LoRA group $\mathbf{G}_{i}^{E}$ and $\mathbf{G}_{i}^{R}$, and adaptively determine the rank of each LoRA. 
Then, the embeddings of all entities $\mathbf{E}_{all}$ and all relations $\mathbf{R}_{all}$ can be denoted as follows:
\begin{eqnarray}
    \label{e_all}
     \mathbf{E}_{all} = {\rm{concat}}(\mathbf{E}_{origin}, \{\mathbf{G}_{j}^{\mathbf{E}} | 1 \leq j \leq i \}) \\
     \label{r_all}
     \mathbf{R}_{all} = {\rm{concat}}(\mathbf{R}_{origin}, \{\mathbf{G}_{j}^{\mathbf{R}} | 1 \leq j \leq i \})
\end{eqnarray}
where $\mathbf{E}_{origin}$ and $\mathbf{R}_{origin}$ denote the origin embeddings of entities and relations in snapshot 0, respectively. 
Finally, we train all LoRAs in $\mathbf{G}_{i}^{\mathbf{E}}$ and $\mathbf{G}_{i}^{\mathbf{R}}$ with new triples $\Delta \mathcal{T}$. 
We take TransE~\cite{bordes2013translating} as the base KGE model, and the loss can be denoted as follows: 
\begin{equation}
    \mathcal{L} = \sum_{(h, r, t) \in \Delta \mathcal{T}_{i}} max(0, f(h, r, t) - f(h', r, t') + \gamma)
    \label{loss}
\end{equation}
where $(h', r, t')$ is the negative triple of $(h, r, t) \in \Delta \mathcal{T}_{i}$, and $f(h, r, t) = |\mathbf{h} + \mathbf{r} - \mathbf{t}|_{L1/L2}$ is the score function of TransE. 
$\mathbf{h} \in \mathbf{E}_{all}$, $\mathbf{r} \in \mathbf{R}_{all}$, and $\mathbf{t} \in \mathbf{E}_{all}$ denote embeddings of $h$, $r$, and $t$, respectively. 
We only train the parameters in $\mathbf{G}_{i}^{\mathbf{E}}$ and $\mathbf{G}_{i}^{\mathbf{R}}$, and fix parameters in all other LoRA groups and origin embeddings. 
Finally, all representations of entities and relations, i.e., $\mathbf{E}_{all}$ and $\mathbf{R}_{all}$ are obtained.

\subsection{Link Predicting}
In the stages of link predicting, we compose all LoRA groups as Equation~\ref{e_all} and~\ref{r_all}. 
Taking link prediction as an example, we freeze all parameters of LoRA groups and initial embeddings. 
For a given query $(h, r, ?)$, we calculate the tail entity $t$ that gives the highest triple score of TransE as the prediction result. 
Notably, \model\ only involves assembling LoRAs into a comprehensive embedding module before the inference stage without requiring additional operations, resulting in no additional time consumption in inference.

\begin{table*}[htb!]
\centering
\setlength{\tabcolsep}{0.9mm}
\begin{tabular}{lcccccccccccccccc}
\hline
\multirow{3}{*}{Dataset} & \multicolumn{3}{c}{Snapshot 0} & \multicolumn{3}{c}{Snapshot 1} & \multicolumn{3}{c}{Snapshot 2} & \multicolumn{3}{c}{Snapshot 3} & \multicolumn{3}{c}{Snapshot 4} \\
 & $N_{E}$ & $N_{R}$ & $N_{T}$ & $N_{E}$ & $N_{R}$ & $N_{T}$ & $N_{E}$ & $N_{R}$ & $N_{T}$ & $N_{E}$ & $N_{R}$ & $N_{T}$ & $N_{E}$ & $N_{R}$ & $N_{T}$ \\
\hline
ENTITY & 2,909 & 233 & 46,388 & 5,817 & 236 & 72,111 & 8,275 & 236 & 73,785 & 11633 & 237 & 70,506 & 14,541 & 237 & 47,326 \\
RELATION & 11,560 & 48 & 98,819 & 13,343 & 96 & 93,535 & 13,754 & 143 & 66,136 & 14,387 & 190 & 30,032 & 14,541 & 237 & 21,594 \\
FACT & 10,513 & 237 & 62,024 & 12,779 & 237 & 62,023 & 13,586 & 237 & 62,023 & 13,894 & 237 & 62,023 & 14,541 & 237 & 62,023 \\
HYBRID & 8,628 & 86 & 57,561 & 10,040 & 102 & 20,873 & 12,779 & 151 & 88,017 & 14,393 & 209 & 103,339 & 14,541 & 237 & 40,326 \\
FB-CKGE & 7505 & 237 & 186,070 & 11,258 & 237 & 31,012 & 13,134 & 237 & 31,012 & 14,072 & 237 & 31012 & 14,541 & 237 & 31010 \\
WN-CKGE & 24,567 & 11 & 55,801 & 28,660 & 11 & 9,300 & 32,754 & 11 & 9,300 & 36,848 & 11 & 9,300 & 40,943 & 11 & 9,302 \\
\hline
\end{tabular}
\caption{The statistics of datasets. $N_{E}$, $N_{R}$ and $N_{T}$ denote the number of cumulative entities, cumulative relations and current triples in each snapshot $i$.}
\label{dataset}
\end{table*}

\section{Experiments}

\subsection{Experimental Setup}
    \subsubsection{Datasets}
We use six datasets in the main experiments: ENTITY, RELATION, FACT, HYBRID, FB-CKGE, and WN-CKGE. 
ENTITY, RELATION, FACT, and HYBRID are traditional datasets for CKGE~\cite{cui2023lifelong}, in which the number of entities, relations, triples and mix of them are growing equally in each snapshot, respectively. 
However, real-world KGs are typically built on a substantial foundational graph, to which a small increment of new knowledge merges in each snapshot. 
To make up for the deficiency of small initial KGs in current CKGE datasets~\cite{hamaguchi2017knowledge,kou2020disentangle,daruna2021continual,cui2023lifelong}, we construct two new datasets for CKGE: FB-CKGE and WN-CKGE, which are based on two widely-used KGE datasets FB15K-237~\cite{dettmers2018convolutional} and WN18RR~\cite{toutanova2015representing}. 
In FB-CKGE and WN-CKGE, we assign 60\% of the total triples to the initial snapshot, and 10\% of the total triples to each next snapshot. 
Compared to traditional datasets ENTITY, RELATION, FACT, and HYBRID based on FB15K-237, FB-CKGE has 2 to 4 times triples in base KGs in snapshot $0$. 
We set the snapshots for all datasets to 5. 
The ratio of train, valid, and test sets for all datasets is 3:1:1. 
The details of the datasets are shown in Table~\ref{dataset}. 
The datasets are available at https://github.com/seukgcode/FastKGE.

\subsubsection{Baselines}
We choose two standard baselines: incremental-parameter fine-tuning methods and full-parameter fine-tuning methods. 
For incremental-parameter fine-tuning methods, we choose PNN~\cite{rusu2016progressive} and CWR~\cite{lomonaco2017core50}. 
For full-parameter fine-tuning methods, we choose replay-based methods GEM~\cite{lopez2017gradient}, EMR~\cite{wang2019sentence}, DiCGRL~\cite{kou2020disentangle}, and regularization-based methods SI~\cite{zenke2017continual}, EWC~\cite{kirkpatrick2017overcoming}, LKGE~\cite{cui2023lifelong}.

\subsubsection{Metrics}
We assess the performance of our model in the link prediction task. 
Specifically, we substitute the head or tail entity of the triples in the test set with all other entities and then calculate and rank the scores for each triple. 
Subsequently, we measure the Mean Reciprocal Rank (MRR), Hits@1, Hits@3, and Hits@10 as evaluation metrics. 
The higher MRR, Hits@1, Hits@3, and Hits@10 indicate better model performance. 
For each snapshot $i$, we compute the average of the above metrics evaluated on all the test sets of current and previous snapshots as the final metric. 
The main results are derived from the model generated at the final time. 
In addition, we calculate the total training time of all snapshots to evaluate the time efficiency.

\subsubsection{Settings}
All experiments are conducted on the NVIDIA RTX 3090Ti GPU. 
The codes of the experiments are supported by PyTorch~\cite{NEURIPS2019_9015}. 
The number of snapshots is set to 5. 
We choose the batch size from [258, 512, 1024] and the learning rate from [1e-1, 2e-1, 3e-1]. 
We choose the Adam as the optimizer. 
In our experiments, we set the entity base rank of LoRA from the range [10, 50, 100, 150, 200] and the relation base rank to 20. 
Also, we set the number of LoRA layers $N$ from the range [2, 5, 10, 20]. 
We set the embedding size for all methods to 200. 
To fairness, all the results are from the average of random five running times, and the patience of early stopping is 3 for all methods to compare time efficiency.

\begin{table*}[htb!]
\centering
\setlength{\tabcolsep}{1mm}
\begin{tabular}{l|ccccc|ccccc|ccccc}
\hline
\multirow{3}{*}{Method} & \multicolumn{5}{c|}{ENTITY} & \multicolumn{5}{c|}{RELATION} & \multicolumn{5}{c}{FACT} \\
 & MRR & H@1 &H@3 & H@10 & T(s) & MRR & H@1 & H@3 & H@10 & T(s) & MRR & H@1 & H@3 & H@10 & T(s) \\
\hline
PNN & 0.229 & 0.130 & 0.265 & \underline{0.425} & 1,979 & 0.167 & 0.096 & 0.191 & 0.305 & 2,186 & 0.157 & 0.084 & 0.188 & 0.290 & 1,511 \\
CWR & 0.088 & 0.028 & 0.114 & 0.202 & 2,218 & 0.021 & 0.010 & 0.024 & 0.043 & 1,828 & 0.083 & 0.030 & 0.095 & 0.192 & 2,807 \\
\hline
GEM & 0.165 & 0.085 & 0.188 & 0.321 & 1,990 & 0.093 & 0.040 & 0.106 & 0.196 & 1,410 & 0.175 & 0.092 & 0.196 & 0.345 & 1,154 \\
EMR & 0.171 & 0.090 & 0.195 & 0.330 & 3,341 & 0.111 & 0.052 & 0.126 & 0.225 & 2,696 & 0.171 & 0.090 & 0.191 & 0.337 & 1,619\\
DiCGRL & 0.107 & 0.057 & 0.110 & 0.211 & 2,416 & 0.133 & 0.079 & 0.147 & 0.241 & 2,481 & 0.162 & 0.084 & 0.189 & 0.320 & 2,204\\
\hline
SI & 0.154 & 0.072 & 0.179 & 0.311 & \underline{1,443} & 0.113 & 0.055 & 0.131 & 0.224 & 1,540 & 0.172 & 0.088 & 0.194 & 0.343 & 1,165 \\
EWC & 0.229 & 0.130 & 0.264 & 0.423 & 2,119 & 0.165 & 0.093 & 0.190 & 0.306 & 1,957 & 0.201 & 0.113 & 0.229 & 0.382 & 1,168 \\
LKGE & \underline{0.234} & \underline{0.136} & \underline{0.269} & \underline{0.425} & 1,515 & \textbf{0.192} & 0.106 & \textbf{0.219} & \textbf{0.366} & \underline{1,242} & \textbf{0.210} & \textbf{0.122} & \textbf{0.238} & \textbf{0.387} & \underline{958} \\
\hline
\textbf{\model} & \textbf{0.239} & \textbf{0.146} & \textbf{0.271} & \textbf{0.427} & \textbf{918} & \underline{0.185} & \textbf{0.107} & \underline{0.213} & \underline{0.359} & \textbf{634} & \underline{0.203} & \underline{0.117} & \underline{0.231} & \underline{0.384} & \textbf{630} \\
\hline
\end{tabular}
\caption{Main experimental results on ENTITY, RELATION and FACT. T denotes the total training time for all snapshots. The bold scores indicate the best results and underlined scores indicate the second best results.}
\label{main_results_1}
\end{table*}

\begin{table*}[htb!]
\centering
\setlength{\tabcolsep}{1mm}
\begin{tabular}{l|ccccc|ccccc|ccccc}
\hline
\multirow{3}{*}{Method} & \multicolumn{5}{c|}{HYBRID} & \multicolumn{5}{c|}{FB-CKGE} & \multicolumn{5}{c}{WN-CKGE} \\
 & MRR & H@1 &H@3 & H@10 & T(s) & MRR & H@1 & H@3 & H@10 & T(s) & MRR & H@1 & H@3 & H@10 & T(s) \\
\hline
PNN & 0.185 & 0.101 & 0.216 & 0.349 & 2,098 & 0.215 & 0.122 & \underline{0.245} & \underline{0.403} & 1,209 & 0.134 & 0.002 & 0.241 & 0.342 & 1,291 \\
CWR & 0.037 & 0.015 & 0.044 & 0.077 & 2,030 & 0.075 & 0.011 & 0.105 & 0.192 & 2,046 & 0.005 & 0.002 & 0.007 & 0.012 & 1,185 \\
\hline
GEM & 0.136 & 0.070 & 0.152 & 0.263 & 2,022 & 0.188 & 0.103 & 0.212 & 0.359 & 1,031 & 0.119 & 0.002 & 0.215 & 0.297 & \underline{951} \\
EMR & 0.141 & 0.073 & 0.157 & 0.267 & 2,976 & 0.180 & 0.097 & 0.204 & 0.346 & 1,500 & 0.114 & 0.002 & 0.205 & 0.286 & 1,153 \\
DiCGRL & 0.149 & 0.083 & 0.168 & 0.277 & 2,231 & 0.149 & 0.091 & 0.160 & 0.261 & 1,742 & 0.057 & 0.001 & 0.155 & 0.166 & 1,012 \\
\hline
SI & 0.111 & 0.049 & 0.126 & 0.229 & 1,300 & 0.187 & 0.102 & 0.211 & 0.359 & 1,088 & 0.115 & 0.001 & 0.209 & 0.289 & 1,077 \\
EWC & 0.186 & 0.102 & 0.214 & 0.350 & 1,535 & \underline{0.218} & \underline{0.124} & 0.247 & 0.410 & 1,196 & 0.136 & 0.003 & 0.248 & 0.338 & 1,013 \\
LKGE & \underline{0.207} & \underline{0.121} & \underline{0.235} & \underline{0.379} & \underline{1,083} & 0.208 & 0.113 & 0.238 & \underline{0.403} & \underline{819} & \underline{0.144} & \underline{0.007} & \underline{0.259} & \underline{0.347} & 1,073 \\
\hline
\textbf{\model} & \textbf{0.211} & \textbf{0.128} & \textbf{0.241} & \textbf{0.382} & \textbf{700} & \textbf{0.223} & \textbf{0.131} & \textbf{0.257} & \textbf{0.405} & \textbf{405} & \textbf{0.159} & \textbf{0.015} & \textbf{0.287} & \textbf{0.356} & \textbf{342} \\
\hline
\end{tabular}
\caption{Main experimental results on HYBRID, FB-CKGE and WN-CKGE.}
\label{main_results_2}
\end{table*}

\subsection{Results}
\subsubsection{Main Results}
The main results are shown in Table~\ref{main_results_1} and Table~\ref{main_results_2}. 
Overall, our method \model\ achieves competitive performance compared to the state-of-the-art methods on all datasets. 
Furthermore, it outperforms all other baselines regarding time efficiency, highlighting its superior training speed.

Firstly, \model\ outperforms all other baselines across all datasets regarding training time efficiency. 
Specifically, on the four traditional datasets ENTITY, RELATION, FACT, and HYBRID, \model\ can save 34\%-49\% training time compared to the fastest baselines. 
Notably, on the two newly constructed datasets FB-CKGE and WN-CKGE, \model\ can save 51\%-68\% training time compared to the fastest baselines. 
This demonstrates that our method with low-rank adapters will be more efficient in larger initial graphs.

Secondly, \model\ has strong competitiveness in performance compared to the best baselines. 
Specifically, \model\ achieves the best performance on two traditional datasets (ENTITY and HYBRID), and two newly constructed datasets (FB-CKGE, and WN-CKGE). 
Compared to the best baselines, \model\ achieves 0.4\%-1.5\%, 0.7\%-1.8\%, 0.2\%-2.8\%, and 0.2\%-0.9\% higher in MRR, Hits@1, Hits@3, and Hits@10, respectively. 
Notably, on two newly constructed datasets (FB-CKGE and WN-CKGE) with extensive initial triples, \model\ can significantly improve performance by 1.5\%, 1.3\%, 2.4\%, and 0.6\% in MRR, Hits@1, Hits@3, and Hits@10 on average compared to the best methods, respectively. 
This proves that \model\ performs better on KGs with larger initial scales, which is more in line with the changes in real life. 
Besides, \model\ also demonstrates strong competitiveness on two other traditional datasets (RELATION and FACT), with only 0.7\%, 0.2\%, 0.6\%, and 0.5\% decrease compared to the optimal baselines in MRR, Hits@1, Hits@3, and Hits@10 on average, respectively. 
This performance decline on RELATION and FACT can be attributed to the limited number of entities they contain. 
Since our method addresses this issue by employing low-rank decomposition and knowledge storage for representing new entities. 
Therefore, the small changes in entity number on RELATION and FACT pose a challenge in showcasing the benefits of our method.

\begin{table*}[htb!]
\centering
\setlength{\tabcolsep}{1mm}
\begin{tabular}{l|ccccc|ccccc|ccccc}
\hline
\multirow{3}{*}{Method} & \multicolumn{5}{c|}{ENTITY} & \multicolumn{5}{c|}{RELATION} & \multicolumn{5}{c}{FACT} \\
 & MRR & H@1 &H@3 & H@10 & T(s) & MRR & H@1 & H@3 & H@10 & T(s) & MRR & H@1 & H@3 & H@10 & T(s) \\
\hline
\model & 0.239 & 0.146 & 0.271 & 0.427 & 918 & 0.185 & 0.107 & 0.213 & 0.359 & 634 & 0.203 & 0.117 & 0.231 & 0.384 & 630 \\
w/o \mec & 0.166 & 0.086 & 0.187 & 0.323 & 1,918 & 0.094 & 0.039 & 0.109 & 0.201 & 1,208 & 0.175 & 0.091 & 0.198 & 0.345 & 979 \\
w/o GL & 0.235 & 0.144 & 0.267 & 0.411 & 755 & 0.168 & 0.099 & 0.196 & 0.295 & 541 & 0.172 & 0.107 & 0.196 & 0.292 & 578 \\
\hline
\end{tabular}
\caption{Ablation results on ENTITY, RELATION and FACT. GL denotes graph layering.}
\label{ablation_results_1}
\end{table*}

\begin{table*}[htb!]
\centering
\setlength{\tabcolsep}{1mm}
\begin{tabular}{l|ccccc|ccccc|ccccc}
\hline
\multirow{3}{*}{Method} & \multicolumn{5}{c|}{HYBRID} & \multicolumn{5}{c|}{FB-CKGE} & \multicolumn{5}{c}{WN-CKGE} \\
 & MRR & H@1 &H@3 & H@10 & T(s) & MRR & H@1 & H@3 & H@10 & T(s) & MRR & H@1 & H@3 & H@10 & T(s) \\
\hline
\model & 0.211 & 0.128 & 0.241 & 0.382 & 700 & 0.223 & 0.131 & 0.257 & 0.405 & 405 & 0.159 & 0.015 & 0.287 & 0.356 & 342 \\
w/o \mec & 0.139 & 0.072 & 0.155 & 0.267 & 1,828 & 0.185 & 0.101 & 0.210 & 0.353 & 1,168 & 0.119 & 0.002 & 0.215 & 0.298 & 959 \\
w/o GL & 0.198 & 0.118 & 0.231 & 0.349 & 674 & 0.218 & 0.126 & 0.252 & 0.308 & 367 & 0.152 & 0.012 & 0.281 & 0.349 & 333 \\
\hline
\end{tabular}
\caption{Ablation results on HYBRID, FB-CKGE and WN-CKGE.}
\label{ablation_results_2}
\end{table*}

\subsubsection{Ablation Results}
This section investigates the effectiveness of incremental low-rank adapter (\mec) learning and graph layering strategy (GL). 
The results are shown in Table~\ref{ablation_results_1} and Table~\ref{ablation_results_2}. 
Firstly, we find that if we remove \mec\ and fine-tune the KGE model directly, the model performance will significantly decrease by 2.8\%-9.1\%, 1.3\%-6.8\%, 3.3\%-9.4\%, and 2.5\%-13.8\% in MRR, Hits@1, Hits@3, and Hits@10 on all datasets, respectively. 
This proves that \mec\ can effectively preserve learned knowledge with incremental low-rank adapters. 
Meanwhile, with \mec\, the training time will decrease obviously by 36\%-65\% on all datasets. 
This proves that using \mec\ can significantly improve training efficiency with low-rank composition compared to direct fine-tuning. 
Secondly, if we remove GL and use one LoRA for all entities, the performance will decrease slightly by 0.4\%-3.1\% on all datasets. 
This proves that our multi-layer LoRAs are more effective than one LoRA with adaptive rank allocation. 
Also, the training time will decrease by 3\%-18\%, indicating that training and assembling multiple LoRAs require additional time. 
Notably, the additional training time only makes the 3\%-9\% time of the total training time on FB-CKGE and WN-CKGE with more initial triples, where the additional time is much smaller than the whole training time.

\subsubsection{Effectiveness of Base Ranks}
\begin{figure}[tb!]
\includegraphics[width=0.48\textwidth]{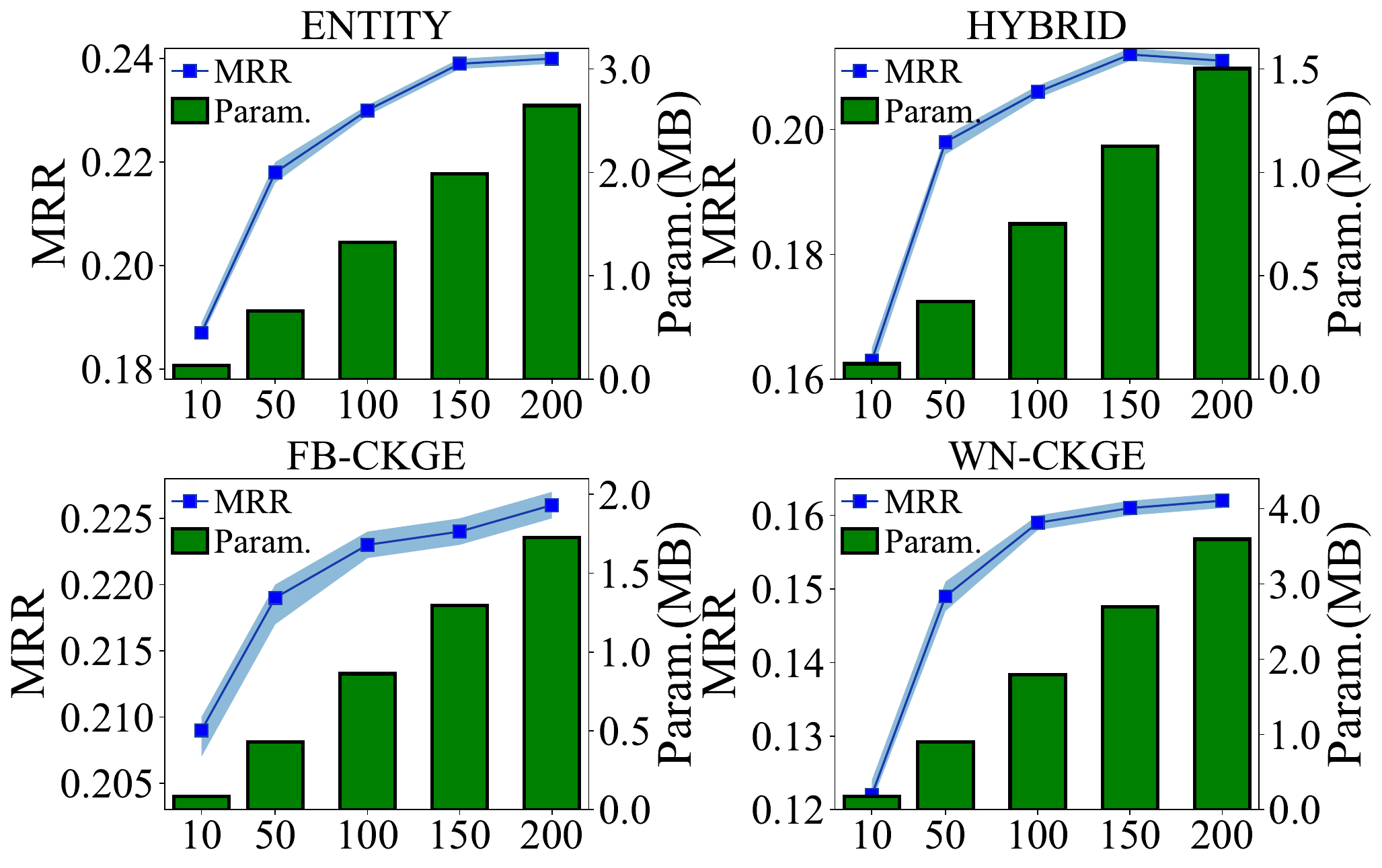}
\caption{Effectiveness of different base ranks from 10 to 200.}
\label{p1}
\end{figure}

This section investigates the effectiveness of different base ranks on the model performance and newly added parameters. 
Since the number of new entities is much more significant than new relations, the number of total embedding parameters is linear to the number of entities. 
Therefore, we only analyze the rank $r_{base}$ for entities. 
We conduct our experiments on ENTITY, HYBRID, FB-CKGE, and WN-CKGE. 
The results are shown in Figure~\ref{p1}. 
Firstly, we find a significant increase in model performance from 1.5\% to 5.2\% in MRR when the rank changes from 10 to 150 in all datasets. 
However, when the rank size exceeds 150, there is only a slight improvement from 0.1\% to 0.2\% in MRR on ENTITY, FB-CKGE, and WN-CKGE. 
Notably, on HYBRID, there is a slight decrease of 0.1\% in MRR when the rank changes from 150 to 200. 
This demonstrates that when the base rank is small, the larger the base rank, the better the performance. 
When the size of the base rank reaches an upper limit, the model performance hardly improves. 
Secondly, we find that the number of newly added parameters in the model increases continuously with the increase of rank. 
Therefore, the model performance does not continually improve with the increase of parameter quantity, but there is a specific upper limit. 
Combining Equation~\ref{condition_2}, we find that when the number of new entities is much greater than the dimension, the upper-rank limit is approximately equal to the dimension, which is consistent with the upper limit of the rank 200 shown in the experimental results.

\subsubsection{Effectiveness of Layer Numbers}
\begin{figure}[tb!]
\includegraphics[width=0.48\textwidth]{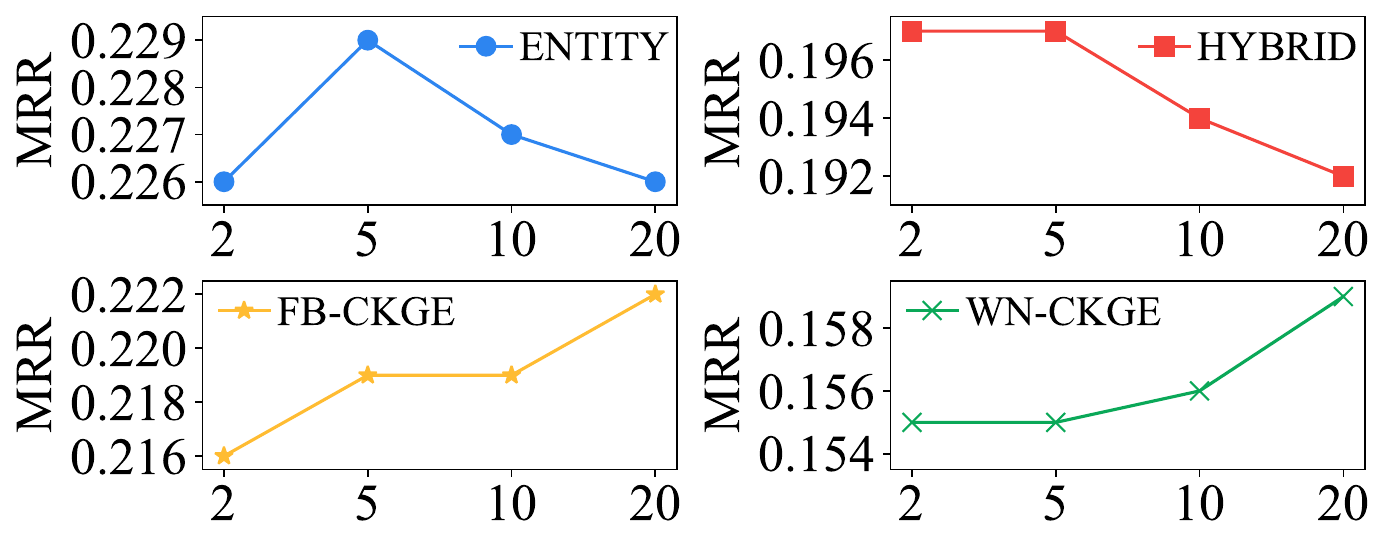}
\caption{Effectiveness of different layer numbers. The horizontal axis represents different layer numbers from 2 to 20.}
\label{p2}
\end{figure}

In this section, we investigate the effectiveness of different layer numbers $N$ on model performance. 
We conduct experiments on two traditional datasets (ENTITY and HYBRID) and two new datasets (FB-CKGE and WN-CKGE), as shown in Figure~\ref{p2}. 
Firstly, on ENTITY and HYBRID, the model performance peaks when layer number equals five and then decreases by 0.4\%-0.5\% in MRR from 5 to 20 layers. 
This indicates that the layer should be set to a small number for the datasets with small initial graphs. 
Secondly, we observe that on FB-CKGE and WN-CKGE, the model performance increases by 0.4\%-0.6\% in MRR when the layer number ranges from 2 to 20. 
It proves that graph layering can achieve more significant results with larger initial KGs.

\subsubsection{Different Base Models}
\begin{figure}[tb!]
\includegraphics[width=0.48\textwidth]{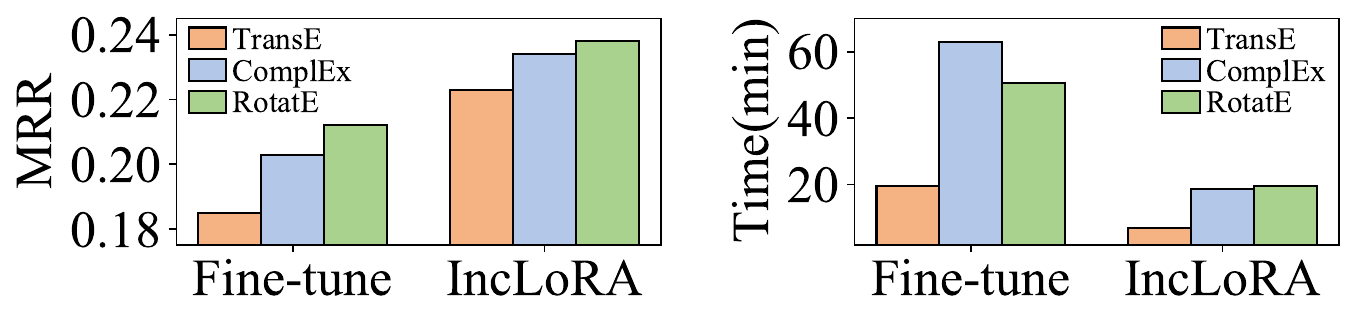}
\caption{Effectiveness of different base KGE models in FB-CKGE.}
\label{p3}
\end{figure}

In this paper, although we choose the most typical model, TransE, as the base KGE model, our method can still be extended easily to other base KGE models. 
To verify the scalability of our method \model, we extend \model\ to two other different types of KGE models, a standard bilinear-based model ComplEx~\cite{trouillon2016complex} and a typical roto-translation-based Model RotatE~\cite{sun2019rotate}. 
We conduct the experiments on FB-CKGE, and the results are shown in Figure~\ref{p3}. 
Firstly, we observe that \model\ outperforms direct fine-tuning by 2.6\%-3.1\% in MRR on ComplEx and RotatE, respectively. 
Meanwhile, for KGE models with better performance, applying our method will also lead to better performance. 
This indicates that \model\ can effectively store the learned knowledge for different KGEs, alleviating catastrophic forgetting in continual learning. 
Secondly, our method can significantly shorten the time from 61.3\% to 70.1\% for different KGEs. 
This demonstrates that \model\ has robust scalability for different KGE models to accelerate the training process.

\begin{table}[t]
\setlength\tabcolsep{0.9pt}
\centering
\begin{tabular}{c|c}
\hline
New entity: \textit{Sedona} & New entity: \textit{Robert Morse} \\
\hline
($f_{dc}$=\textcolor[RGB]{45,133,240}{4.7e-4}, dis=1) & ($f_{dc}$=\textcolor[RGB]{45,133,240}{1.18e-3}, dis=1) \\
\hline
\hline
\multicolumn{2}{c}{Queries} \\
\hline
\hline
(Michelle Branch, location, ?) & (Mad Man, award\_winner, ?) \\
\hline
\hline
\multicolumn{2}{c}{Prediction Results of \model} \\
\hline
\hline
$1st:$ \textit{Sedona} & $1st:$ \textit{Robert Morse} \\
$2nd:$ Orlando & $2nd:$ Bryan Batt \\ 
$3rd:$ Detroit & $3rd:$ Jared Harris \\
\hline
\hline
\multicolumn{2}{c}{Prediction Results of \model\ without graph layering} \\
\hline
\hline
$1st:$ \textit{Sedona} & $1st:$ Maria Jacquemetton \\
$2nd:$ Wiltshire & $2nd:$ Mathew Weiner \\
$3rd:$ Cornwall & $3rd:$ \textit{Robert Morse} \\
\hline
\end{tabular}
\caption{Case study for the graph layering strategy.}
\label{case_study}
\end{table}

\subsection{Case Study}
We conduct a case study to verify that \model\ learns more critical entities well by graph layering with dynamic rank allocation. 
As shown in Table~\ref{case_study}, we select two new entities with different importance, namely, with the same distance from the old graph (dis=1) but different $f_{dc}$, and test the prediction results with layering and non-layering strategies. 
Results show that for entity \textit{Sedona} with small $f_{dc}=4.7e-4$, both strategies can rank it first. 
However, for entity \textit{Robert Morse} with larger $f_{dc}=1.18e-3$, the rank of the correct answer will increase after layering. 
It proves that graph layering learns critical new entities effectively without affecting others.

\section{Conclusion}
This paper proposes a novel fast learning framework for CKGE, \model, which utilizes an \mec\ mechanism to preserve learned knowledge well and accelerate fine-tuning. 
Firstly, to alleviate catastrophic forgetting, we conduct graph layering for new knowledge to achieve separate storage of old and new knowledge. 
Moreover, to reduce training costs, we propose incremental low-rank adapter learning to learn new knowledge efficiently with adaptive rank allocation. 
In the future, we will explore how to conduct CKGE learning when the knowledge of growing KGs is forgotten or modified.

\section*{Acknowledgments}
We thank the reviewers for their insightful comments. 
This work was supported by National Science Foundation of China (Grant Nos.62376057) and the Start-up Research Fund of Southeast University (RF1028623234). 
All opinions are of the authors and do not reflect the view of sponsors.

\bibliographystyle{named}
\bibliography{ijcai24}

\end{document}